\documentclass[10pt,twocolumn,letterpaper]{article}

\usepackage{cvpr}
\usepackage{times}
\usepackage{epsfig}
\usepackage{graphicx}
\usepackage{amsmath}
\usepackage{amssymb}
\usepackage{subfig}
\usepackage{epstopdf}

\usepackage[breaklinks=true,bookmarks=false]{hyperref}

\cvprfinalcopy % *** Uncomment this line for the final submission

\def\cvprPaperID{****} % *** Enter the CVPR Paper ID here

\begin{document}

%%%%%%%%% TITLE
\title{ZSTAD: Zero-Shot Temporal Activity Detection}

\author{Lingling Zhang$^{1,2}$, Xiaojun Chang$^3$, Jun Liu$^{1,4}$, Minnan Luo$^{1,4}$, Sen Wang$^{5}$,\\ Zongyuan Ge$^{3}$, Alexander Hauptmann$^{6}$\\
$^1$School of Computer Science and Technology, Xi’an Jiaotong University, Xian, China\\
$^2$Ministry of Education Key Lab For Intelligent Networks and Network Security, Xian, China\\
$^3$Faculty of Information Technology, Monash University, Australia\\
$^4$National Engineering Lab for Big Data Analytics, Xi’an Jiaotong University, Xian, China \\
$^5$School of Information Technology and Electrical Engineering, The University of Queensland, Australia\\
$^6$School of Computer Science, Carnegie Mellon University, USA\\
{\tt\small zhanglingling@stu.xjtu.edu.cn, cxj273@gmail.com, \{liukeen, minnluo\}@xjtu.edu.cn,} \\
{\tt\small sen.wang@uq.edu.au, zongyuan.ge@monash.edu, alex@cs.cmu.edu}\\
% For a paper whose authors are all at the same institution,
% omit the following lines up until the closing ``}''.
% Additional authors and addresses can be added with ``\and'',
% just like the second author.
% To save space, use either the email address or home page, not both
}

\maketitle
%\thispagestyle{empty}

%%%%%%%%% ABSTRACT
\begin{abstract}
	An integral part of video analysis and surveillance is temporal activity detection, which means to simultaneously recognize and localize activities in long untrimmed videos.
	Currently, the most effective methods of temporal activity detection are based on deep learning, and they typically perform very well with large scale annotated videos for training. 
	However, these methods are limited in real applications due to the unavailable videos about certain activity classes and the time-consuming data annotation.
	To solve this challenging problem, we propose a novel task setting called zero-shot temporal activity detection (ZSTAD), where activities that have never been seen in training can still be detected. 
	We design an end-to-end deep network based on R-C3D as the architecture for this solution. 
	The proposed network is optimized with an innovative loss function that considers the embeddings of activity labels and their super-classes while learning the common semantics of seen and unseen activities. 
	Experiments on both the THUMOS’14 and the Charades datasets show promising performance in terms of detecting unseen activities. 	
\end{abstract}

%%%%%%%%% BODY TEXT
\section{Introduction}
Given its importance to video analysis and surveillance, temporal activity detection is one of the most studied tasks in computer vision \cite{shou2017cdc}. 
Most videos are untrimmed and only contain a few interesting activities among a long stream of nondescript scenes. 
Hence, the goal of temporal activity detection is to simultaneously recognize and categorize specific activities in a video along with their start and end times \cite{shou2018autoloc, huang2018sap}.
As with many other tasks, deep learning has led to a step-change in the speed and accuracy of temporal activity detection, as demonstrated in studies like \cite{zhao2017temporal,gao2017turn,gao2017cascaded,xu2017r}.
However, these deep methods rely heavily on a fully-supervised training scheme. 
Long videos where every activity class is already annotated are rare, and manual annotation is expensive and time-consuming, which means an alternative approach, like weakly-supervised \cite{wang2017untrimmednets,paul2018w}, semi-supervised or unsupervised learning, is needed.

\begin{figure*}[t]
	\centering
	\includegraphics[width=6.4 in]{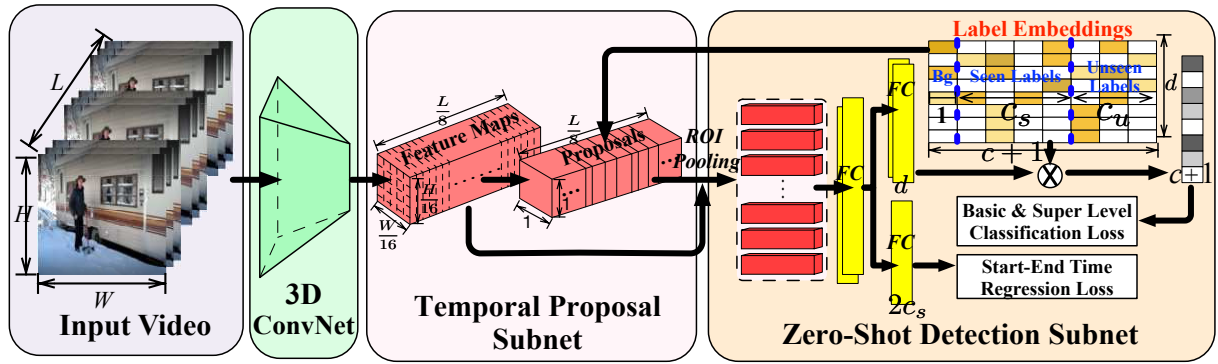}
	\vspace{-0.5 em}
	\caption{Diagram of zero-shot temporal activity detection network. 
		Purple part: the input videos are resized into fixed dimensionality;
		Green part: 3D ConNet extracts the deep features for input videos;
		Pink part: Temporal proposal subnet (TPN) generates the proposals for seen and unseen activities;
		Orange part: Zero-shot detection subnet (ZSDN) detects the seen and unseen activities by utilizing their label embeddings.
		\label{fig:frame}}
	\vspace{-0.6 em}
\end{figure*}

In recent years, zero-shot learning (ZSL) has emerged as a novel approach to this very problem. 
ZSL can be thought of as a special case of supervised learning, where the training and testing classes (\emph{i.e.}, the seen and unseen classes) are disjoint but semantically related \cite{palatucci2009zero, rahman2018unified}.
In fact, generally speaking, the training dataset does not even contain any instances labeled with testing classes \cite{meng2019zero, luo2017zero}. 
ZSL works by exploiting useful prior knowledge, such as common attributes or label embeddings, and then transfers the learned knowledge from seen classes to unseen classes. 
So far, studies on ZSL have been dominated by zero-shot recognition (ZSR), \emph{i.e.}, classification problems. 
These approaches are often based on the assumption that each sample only contains a single dominant example from one of the classes (whether seen or unseen). 
Hence, the overall aim is to classify each unseen sample to an unseen class in the testing stage. 
On this proviso, ZSR would not be directly applicable to temporal activity detection for two reasons:
(1) Temporal activity detection is of most value with long videos, which means the sample would almost certainly contain multiple activities. 
(2) Part of temporal activity detection is predicting the start and end times for each activity in addition to recognizing the class of behavior. 
Thus, in this paper, we present a novel task setting for ZSR, called zero-shot temporal activity detection (ZSTAD), that is designed specifically for long untrimmed videos and can also predict the temporal information surrounding those activities.

Unlike simple ZSR, ZSTAD not only recognizes activities that are not seen during training, but it also localizes the start and end times of the activities. 
ZSTAD is implemented in a novel end-to-end deep network, as shown in Figure \ref{fig:frame}. 
The network uses two types of prior information to mine the common semantics of seen and unseen activities: label embeddings from Word2Vec \cite{goldberg2014word2vec} and the super-classes of those labels. 
For ZSTAD, the label embedding for the background class is crucial, however, it does not directly correspond to any embeddings in Word2Vec.
Hence we solve an optimization problem to obtain a special vector for the background.
Note that this learned embedding is far away from all the activity label embeddings in the semantic space. 
With this prior information, the ZSTAD deep network thus consists of four parts: 
(1) the input process, which resizes the original video into a sequence of RGB frames with fixed dimensionality; 
(2) a 3D convolutional subnet (3D ConvNet), which extracts the deep features from the input videos; 
(3) a temporal proposal subnet (TPN), which uses the learned deep features and the embedding of the background class as inputs to generate proposals for seen and unseen activities; 
and (4) a zero-shot detection subnet (ZSDN), which uses the label embeddings of all activities to predict a proposal's class and start/end times. 

To optimize the deep network, we design an innovative loss that combines the classification loss and boundary regression loss of both the TPN and ZSDN subnets. 
Note that the classification loss for ZSDN consists of two terms: the basic-class classification loss and the super-class clustering loss. 
The basic-class classification loss guarantees that the score for the true label is the largest of all labels. 
The super-class clustering loss controls the labels that are similar to the true label, to obtain the higher scores than other labels. 
Four contributions of the paper are summarized as follows: 
\begin{itemize}
	\vspace{-0.3 em}
	\item We propose a new task setting called ZSTAD to simultaneously classify and localize the activities in long untrimmed videos, even if the activity classes have never been seen during training.
	\vspace{-0.6 em}
	\item  A novel end-to-end deep network is developed for ZSTAD task, which uses prior information from label embeddings of the background, seen activities, and unseen activities. The background label embedding is derived by solving an optimization problem and is far away from the other label embeddings. 
	\vspace{-0.6 em}
	\item We design an innovative classification loss for ZSDN subnet, which 
	integrates the basic-class classification loss and super-class clustering loss together. 
	Note that the super-class prior information carries the semantic correlation among activity labels.	
	\vspace{-0.6 em}	
	\item We conduct extensive experiments on two widely-used video datasets, THUMOS'14 and Charades, to evaluate the effectiveness and superiority of the proposed deep model. 
\end{itemize}

\section{Related Work}
\subsection{Temporal Activity Detection}
The literature on temporal activity detection can be divided into two groups according to how the activity proposals are generated. 
In the first category, activity classifiers are applied to sliding windows generated by scanning across the video. 
For example, Wang \emph{et al.} \cite{wang2014action} used a temporal sliding window to generate video clips and then extracted motion and appearance features for action classification. 
Shou \emph{et al.} \cite{shou2016temporal} generated segments with varied lengths via sliding windows, then fed them into 3D ConvNets for action recognition. 
And Gaidon \emph{et al.} \cite{gaidon2013temporal} described a sliding central frame based on a generative model of temporal structures. 
These sliding window methods not only have a high computation cost due to the large number of windows, but also constrain the boundary of detected activity segments to some extent \cite{xu2018similarity}.

To avoid these issues, the second category of methods detects activities with arbitrary lengths by modeling the temporal evolution of the video. 
For instance, Gao \emph{et al.} \cite{gao2017turn} presented an effective temporal unit regression network for generating temporal action proposals. 
Xiong \emph{et al.} \cite{xiong2017pursuit} introduced a new proposal generation scheme called temporal actionness grouping that can efficiently generate candidates with accurate temporal boundaries. 
Xu \emph{et al.} \cite{xu2017r} put forward the first end-to-end activity detection model, RC3D, by combining the proposal generation and classification stages together. 
%Qiu \emph{et al.} \cite{qiu2018precise} proposed a novel three-phase  temporal proposal framework for action localization, which used multi-stage temporal coordinate regression under various temporal granularities.
Compared to the sliding window methods, these temporal evolution methods have attracted more attention because of their added flexibility in predicting arbitrary activity start and end times.

\subsection{Zero-Shot Learning}
Zero-shot learning (ZSL) is designed to recognize samples of classes that are not seen during training \cite{zhu2018towards, zhang2019triple}. 
The idea is to learn shared knowledge from prior information and then transfer that knowledge from seen classes to unseen classes \cite{niu2019zero,qin2017zero}. 
Common attributes, such as color, shape, and similar properties, are the typical forms of prior information. 
Lampert \emph{et al.} \cite{lampert2014attribute} pre-learned the attribute classifiers independently to accomplish ZSL on unseen classes, while Parikh \emph{et al.} \cite{parikh2011relative} learned relative attributes to enhance ZSL accuracy. 
Attribute-based methods have achieved promising results on ZSL, but they have poor scalability because the attributes need to be manually defined. 
Semantic embeddings of seen and unseen labels, which are another type of prior information, do not have this problem \cite{xu2015semantic}. 
They are generally learned in an unsupervised manner with a method such as Word2Vec \cite{goldberg2014word2vec} or GloVe \cite{pennington2014glove}. 
Socher \emph{et al.} \cite{socher2013zero}, for example, mapped samples and labels into a shared semantic space and introduced a binary variable to indicate whether an instance was in seen or unseen class. 
Zhang \emph{et al.} \cite{zhang2015zero} developed a semantic similarity embedding method by expressing seen and unseen data as a mixture of seen class proportions. 
Compared to the attribute-based methods, label embedding methods are more practical and popular because this type of prior information is easily accessible from open text corpora.

Beyond these studies, which focus on ZSR problems, four notable studies on zero-shot object detection (ZSOD) have recently appeared in the image processing literature. 
Zhu \emph{et al.} \cite{zhu2018zero} presented a novel ZSOD architecture that fused semantic attribute information with visual features to predict the locations of unseen objects. 
Bansal \emph{et al.} \cite{bansal2018zero} proposed a background-aware ZSOD model based on visual-semantic embeddings. 
Demirel \emph{et al.}'s \cite{demirel2018zero} solution for ZSOD aggregated both label embeddings and convex combinations of semantic embeddings. 
Last, Rahman \emph{et al.} \cite{rahman2018zero} has proposed the first end-to-end deep network for ZSOD based on the Faster RCNN \cite{ren2015faster} framework. 

To the best of our knowledge, this paper is the first to apply the idea of ZSL to temporal activity detection.

\section{Zero-Shot Temporal Activity Detection}
\label{Zero-Shot Temporal Activity Detection}
This section begins with a description of the problem setting of ZSTAD. 
We then introduce the detailed architecture of the proposed deep network, and design a novel objective loss to optimize the network's parameters. 

\subsection{Problem Description}
\label{Problem Description}
In the framework of ZSTAD, there are $n$ untrimmed videos $\mathcal{X}=\{\mathbf{x}_1, \mathbf{x}_2, \cdots, \mathbf{x}_n\}$ about $c$ activity classes.
The first $n_s$ videos $\mathcal{X}_s=\{\mathbf{x}_1, \mathbf{x}_2, \cdots, \mathbf{x}_{n_s}\}$ are labeled for training with temporal annotations that cover the first $c_s$ seen activity classes. 
The remaining $n_u$ videos $\mathcal{X}_u=\{\mathbf{x}_{n_s+1}, \mathbf{x}_{n_s+2}, \cdots, \mathbf{x}_{n}\}$ are unlabeled for testing, with each containing at least one activity in $c_u$ unseen classes.
Motivated by previous studies on ZSL, label embeddings generated from an unsupervised method, such as Word2Vec, are still used to measure the semantic relationships in both seen and unseen classes. 
The label embedding set of $c_s$ seen activities is denoted as $\mathcal{L}_s=\{\mathbf{\ell}_1, \mathbf{\ell}_2, \cdots, \mathbf{\ell}_{c_s}\}$, accordingly, $\mathcal{L}_u=\{\mathbf{\ell}_{c_s+1}, \mathbf{\ell}_{c_s+2}, \cdots, \mathbf{\ell}_{c}\}$ for $c_u$ unseen activities.
Especially, the label embedding for background class is not directly available from Word2Vec and, without it, the model cannot determine whether the video contains any activity or not.
Therefore, we propose the following optimization problem to overcome this problem:
\begin{align}
	\label{bg_embedding}
	{\ell}_{bg} = \arg\min_{{\ell}_{bg}} \sum_{j=1}^c max(0, s({\ell}_{bg},\ell_j)-\Delta_{bg})^2,
\end{align}
where $\Delta_{bg}$ is the margin hyper-parameter. 
Solving this problem results in a special vector ${\ell}_{bg} $ that is far away from all embeddings of activity labels.
The function $s({\ell}_{bg},\ell_j)$ measures the semantic similarity between label embeddings ${\ell}_{bg}$ and ${\ell}_j$, which is calculated according to cosine distance. %\emph{i.e} $s({\ell}_{bg},\ell_j) = \frac{\ell_{bg}^\top \ell_j}{\| \ell_{bg} \|\| \ell_{j} \|}$. 
In effect, this equation stipulates that the background label is at most $\Delta_{bg}$ similar to other activity labels in the semantic embedding space. 
The resulting set of all label embeddings is denoted as $\mathcal{L}_{bg} = \{\ell_{bg}, \ell_1,\ell_2,\cdots,\ell_{c}\}$.

Rahman \emph{et al.} \cite{rahman2018zero} proved that super-classes over label embeddings benefit ZSL. 
Therefore, we have incorporated super-classes into the embeddings $\mathcal{L}_{bg} = \{\ell_{bg}, \ell_1,\ell_2,\cdots,\ell_{c}\}$ as another type of prior information in the ZSTAD framework. 
More specifically, we partition $c$ activity label embeddings into $c^+$ disjoint super-classes such that the activity labels in the same super-class share high semantic similarity, while ensuring that the labels from different super-classes are as dissimilar as possible.
These activity super-classes are denoted as $\mathcal{Z} = \{z_1,z_2,\cdots,z_{c^+}\}$, where $z_q$ is a set of label indexes whose corresponding embeddings are partitioned into the $q$-th super-class. 
Namely, $z_q = \{j \in [1,c], s.t., g(\ell_j) = q\}$, where the function $g(\cdot)$ maps each activity label embedding to its corresponding super-class $z_{g(.)}$. 
Practically, the mapping function $g(\cdot)$ could be performed by any number of popular clustering algorithms. %such as the typical c-means \cite{hartigan1979algorithm}, fuzzy c-means \cite{bezdek1984fcm} and spectral clustering \cite{ng2002spectral} methods. 
Further, the background label embedding $\ell_{bg}$ is set into an individual super-class $z_{bg}$ to ensure that it is clearly different from the other activity labels in the semantic embedding space. 
Ultimately, the overall super-classes is represented as $\mathcal{Z}_{bg} = \{z_{bg},z_1,z_2,\cdots,z_{c^+}\}$.

To summarize, the ZSTAD model is trained on labeled videos $\mathcal{X}_s$ with temporal annotations of seen activities, with the objective of recognizing and localizing unseen activities in unlabeled videos $\mathcal{X}_u$. 
The common semantics between seen and unseen activities are reflected in the prior information from the label embeddings $\mathcal{L}_{bg}$ and super-classes $\mathcal{Z}_{bg}$.

\begin{figure}[t]
	\centering
	\includegraphics[width=2.6 in]{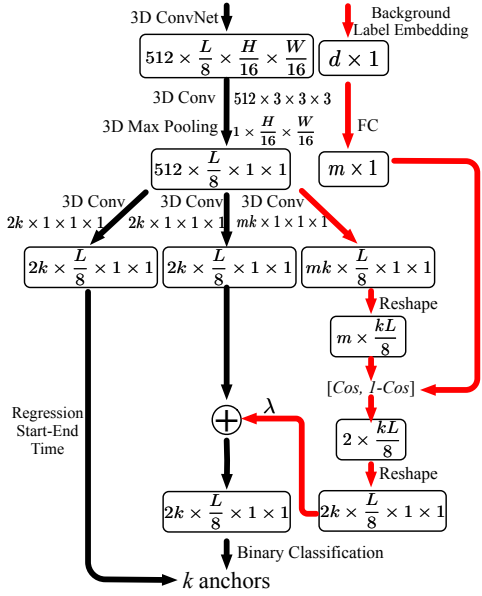}
	\caption{Diagram of the temporal proposal subnet.
		\label{fig:prop_frame}}
	\vspace{-0.6em}
\end{figure}
\subsection{Network Architecture}
The deep network for ZSTAD is illustrated in Figure  \ref{fig:frame}. 
Its backbone is the R-C3D framework \cite{xu2017r} due to the R-C3D's superior performance in temporal activity detection. 
The four colored panels delineate the four essential components. 
The purple panel shows the input process, where videos are transformed into a sequence of RGB frames with dimension $\mathbb{R}^{3\times L \times H \times W}$. 
Here, $L$ denotes the number of frames and $H, W$ represent the height and width of each frame. 
The 3D ConvNet (the green panel) extracts the deep features from the input videos. The architecture of this subnet is similar to that proposed in \cite{tran2015learning}, \emph{i.e.}, eight 3D convolutional layers (conv$1a$ to conv$5b$) and four max-pooling layers (pool$1$ to pool$4$), and the output is a feature map with dimension $\mathbb{R}^{512\times \frac{L}{8} \times \frac{H}{16} \times \frac{W}{16}}$. 
The pink panel houses the TPN, which takes the video's feature map as its input and outputs proposal segments of variable lengths. 
The ZSDN, in the orange panel, recognizes the classes of the generated proposals and fine-tunes their start/end times. 
The two key components of the framework are obviously TPN and ZSDN. 
These are discussed in more detail in the following sections.

\subsubsection{Temporal Proposal Subnet (TPN)}
The architecture of the TPN subnet is illustrated in detail in Figure \ref{fig:prop_frame}. 
This subnet generates high-quality proposals and can correctly distinguish whether the proposal contains an example of any activity. 
Its anchor segments are predefined multiscale windows centered on $L/8$ uniformly distributed temporal locations, where the maximum possible number of proposals for each location is $k$. 
Therefore, the total number of segments for each input video is $kL/8$.  
The output of the 3D ConvNet first passes through a 3D convolutional filter and a max-pooling layer, which is represented as a temporal-only feature map $C_{tpn} \in \mathbb{R}^{512 \times \frac{L}{8} \times 1 \times 1}$.
Namely the 512-dimensional feature is produced for each temporal location.
This feature map is then fed into three modules: a  boundary regression module (left flow), a basic classification module (center flow), and an improved classification module (right flow).
The boundary regression module is used to predict the offset and length of each segment. 
The purpose of two binary classification modules is to score each proposal as either an activity or background. 
Their outputs are added together to jointly determine the final scores. 

Note that the architectures of both the boundary regression module and the basic classification module are the same as R-C3D. 
The improved classification module was designed by us to refine the results produced by the basic module by introducing the background label embeddings. 
The improved classification module contains an additional 3D convolutional layer with $mk$ channels on the feature map $C_{tpn}$. 
The output is then reshaped as $m\times (kL/8)$, \emph{i.e.}, each anchor segment is represented as an $m$-dimensional feature vector. 
The $d$-dimensional background label embedding is also mapped into $m$-dimensional space with a fully-connected layer. 
In this way, the improved classification module determines its binary scores by computing the cosine distance between $m$-dimensional features of the anchor segment and the background label embedding. 
The improved classification module can decrease the mistake of regarding unseen activities as background in the testing stage, because the label embedding of background is far away from that of unseen activities.

\subsubsection{Zero-Shot Detection Subnet (ZSDN)} 
The TPN subnet generates a set of candidate activity proposals by automatically ranking the confidence scores of all anchor segments. 
The non-maximum suppression algorithm is applied first to eliminate the proposals with large overlaps or low confidence scores. 
Next, three-dimensional region of interest (3D RoI) pooling with $1 \times 4 \times 4$ grids maps the selected proposals, which are of different lengths, into a fixed size of $512 \times 1 \times 4 \times 4$. 
Each proposal's pooled feature passes through two fully-connected layers, and then onwards into a zero-shot classification module and a boundary regression module. 
Descriptions follow.
\begin{itemize}
	\vspace{-0.3 em}
	\item The zero-shot classification module includes two fully-connected layers that project the input proposal’s features onto $d$-dimensional semantic space. 
	A score vector $\mathbf{p}_i \in \mathbb{R}^{c+1}$ is calculated for each proposal by computing the cosine distance between the proposal's feature vector and all label embeddings $\mathcal{L}_{bg}$. 
	Vector $\mathbf{p}_i$ consists of the possibilities that the proposal belongs to the background class and the $c$ activity classes.
	\vspace{-0.6 em}
	\item The boundary regression module is one fully-connected layer at the top of the network, which generates $2\times{c_s}$ outputs. These outputs are used to refine each proposal's start/end times.
\end{itemize}

\subsection{Training and Inference}
The network is trained by jointly optimizing the classification and regression losses for both the TPN and ZSDN subnets. 
Note that three losses, including the classification and regression losses of the TPN and the regression loss of the ZSDN, are set as the work in \cite{xu2017r}. 
This last loss, \emph{i.e.}, the classification loss of the ZSDN subnet, is integral to accomplishing the ZSTAD task, which is designed as follows:
\begin{align}
	\label{cls_loss}
	L_{zs-cls} =\frac{1}{N} \sum_i (L_{bc}(\mathbf{p}_i,p^*_i) + \beta L_{sc}(\mathbf{p}_i,p^*_i)),
\end{align}
where $N$ denotes the number of proposals in the ZSDN subnet, and $i$ stands for their index during the training procedure. 
The vector $\mathbf{p}_i \in \mathbb{R}^{c+1}$ is the output of the ZSDN's zero-shot classification module. 
It represents the predicted probability distribution over the background and $c$ activity classes. 
The integer $p^*_i \in \{0,1,\cdots,c\}$ represents the ground-truth label of the $i$-th proposal. 
If $p^*_i = 0$, the proposal does not contain any activities, otherwise it belongs to the $p^*_i$ -th activity class. 
On the whole, the classification loss $\mathcal{L}_{zs-cls}$ consists of two components: a basic-class classification loss $\mathcal{L}_{bc}$ and a super-class clustering loss $\mathcal{L}_{sc}$.
$\beta$ is a hyper-parameter that controls the trade-off between these two terms. 
Given a proposal, the basic-class classification loss guarantees that the predicted score for its true class will be the largest in the $(c+1)$-dimensional score vector $\mathbf{p}_i$.
It is evaluated with a general softmax loss as follows: 
\begin{align}
	\label{cls_bc_loss}
	L_{bc}(\mathbf{p}_i,p^*_i) = -\log{p}_{ip^*_i}.
\end{align}
The $L_{sc}$ clustering loss considers the prior information from the super-classes $\mathcal{Z}_{bg}$ of the activities.
It controls the predicted scores of activity labels in the same super-class as the true label, should be higher than those of other activity labels in different super-classes. 
The value $L_{sc}$ is derived from the following hinge loss:
\begin{align}
	\label{cls_sc_loss}
	L_{sc}(\mathbf{p}_i,p^*_i) = \frac{1}{N_{sc}} \sum_{j_1\notin z_{s_i^*}} \sum_{j_2 \in z_{s_i^*} } \max(0,p_{ij_1}-p_{ij_2} + \Delta_{sc}),
\end{align}
where $\Delta_{sc}$ is a margin hyper-parameter that can be determined through cross-validation. 
The integer $s_i^*$ is the super-class index of the ground-truth activity label embedding $\ell_{p_i^*}$, \emph{i.e.} $s_i^* = g(\ell_{p_i^*})$. 
The number of $N_{sc}$ is equal to $(c+1-|z_{s_i^*}|) \times |z_{s_i^*}|$, where $|z_{s_i^*}|$ denotes the total number of activity labels in the super-class $z_{s_i^*}$. 

Note that no relative time offsets and lengths are determined for unseen activities because no unseen activity samples are included in the training procedure.
Therefore, the testing stage simply produces a $(2\times c_s)$ matrix denoting the two parameterized coordinates for each seen activity class. 
In this case, the relative offset and length of unseen activity are approximated through the coordinates of closely related seen activities. 
This strategy was first proposed by Rahman \emph{et al.} \cite{rahman2018zero}, and it has proven to be effective for zero-shot detection because the visual features of unseen activities are usually similar to those of their close seen activities.

\section{Experiment}
\label{Experiment}
\subsection{Datasets}
\label{datasets}
We conduct experiments with two video datasets: THUMOS'14 and Charades. Their details follow.
\begin{itemize}
	\vspace{-0.3 em}
	\item \textbf{THUMOS'14} \cite{THUMOS14}: 
	This dataset contains 20 activity classes for temporal activity detection and consists of four parts: training data, validation data, testing data, and background data. 
	We use the validation data (200 untrimmed videos) to train our network and the test data (213 untrimmed videos) to evaluate the model's performance. 
	12 activities are selected as seen classes with the remaining 8 activities as the unseen classes.
	\vspace{-0.6 em}
	\item \textbf{Charades} \cite{sigurdsson2016hollywood}: 
	This is another widely-used dataset for activity recognition and detection. 
	It comprises 9848 videos over 157 daily indoor activities collected through Amazon Mechanical Turk. 
	We use 7985 videos for training and 1863 for testing. 120 activities are chosen as seen classes; 37 activities are unseen.
\end{itemize}
As described in Section \ref{Problem Description}, the two types of prior information used are the activity label embeddings and their super-classes. 
The labels in THUMOS'14 consist of one or two nouns, so we use the average of their 300-dimensional features from Word2Vec for the activity label embeddings. 
The labels in Charades are gerund phrases, such as ``Taking a picture of something''. 
To ensure the accuracy of label embeddings, we remove the quantifiers and prepositions from these phases, then represent the remaining words as mean Word2Vec features.
The background label embedding is obtained by solving the optimization problem Eq. (\ref{bg_embedding}) with a parameter setting of $\Delta_{bg} = 0.1$. 
%The background label embedding relates to all the activity labels in a specific dataset and is obtained by solving the optimization problem Eq. (\ref{bg_embedding}) with a parameter setting of $\Delta_{bg} = 0.1$. 
The resulting activity label embeddings are then grouped into several super-classes with the self-tuning spectral clustering algorithm \cite{zelnik2005self}. 
The background label is assigned into a individual super class because it is far away from other labels.
\begin{figure}[t]
	\centering
	\subfloat[THUMOS'14 \label{fig:result_thumos14} ]{\includegraphics[width=1.0\linewidth]{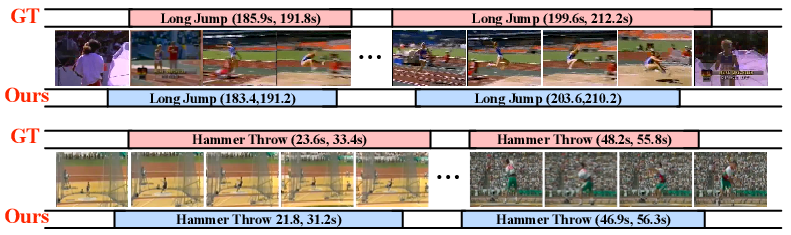}} \\
	\vspace*{-0.5em}
	\subfloat[Charades \label{fig:result_charades}]{\includegraphics[width=1.0\linewidth]{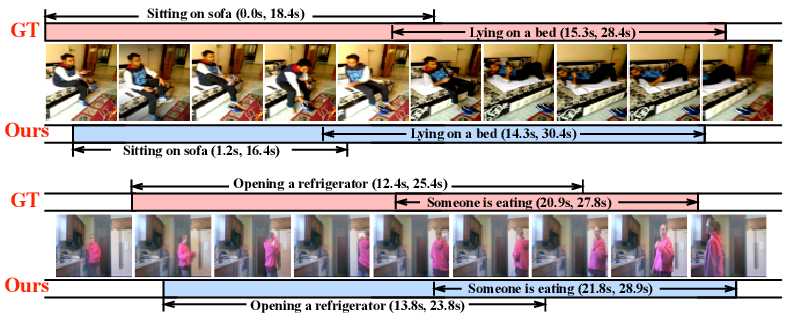}}
	\caption{Some prediction results of zero-shot temporal activity detection over two datasets. \label{fig:result}}
	\vspace{-0.6em}
\end{figure}

\subsection{Experimental Setup}
\label{experimantal setup}
To limit the GPU memory consumed, the network input with the THUMOS'14 dataset is set to a sequence of RGB frames with a dimension of $3 \times 512 \times 112 \times 112$, \emph{i.e.}, the equivalent of a 30-second video clip. 
The input with the Charades dataset is limited to $3 \times 768 \times 112 \times 112$, which is about 154 seconds.
Note that the ZSL setting demands that the training dataset does not contain any instances labeled with testing classes. 
Therefore, we remove any clips containing unseen activities from the training video clips. 
In addition, we ensure that each clip in the testing set contains at least one unseen activity.
We build the network in the open-source Caffe framework \cite{jia2014caffe}, and the parameters of the 3D ConvNet is pre-trained on the Sports-1M dataset \cite{KarpathyCVPR14} to avoid overfitting. 
The number of anchor segments $k$ are set to values within [2,4,5,6,8,9,10,12,14,16] for THUMOS'14 and within [1,2,3,4,5,6,7,8,10,12,14,16,20,24,28,32,40,48] for Charades. 
The three hyper-parameters are set as follows: 
the trade-off in the improved TPN subnet is set to $\lambda=0.6$; the trade-off in zero-shot classification loss is set to $\beta=0.1$ (Eq. \ref{cls_loss}); and the margin $\Delta_{sc}$ in the clustering loss function is set from [0.05,0.30] for the THUMOS'14  dataset and from [0.03,0.15] for Charades (Eq. (\ref{cls_sc_loss})). 
Training is conducted with stochastic gradient descent at a learning rate of $0.0001$, a momentum of $0.9$, and a weight decay of $0.00005$ to optimize the proposed deep ZSTAD network.
\begin{table}[t]
	\center
	\tabcolsep=2.05 pt
	\caption{Zero-shot temporal activity detection results on THUMOS'14 w.r.t mAP (\%) at different IoU thresholds. \label{tab:thumos14_1}}
	{
		\begin{tabular}{c|ccccccc}
			\hline
			&$\alpha$=$0.1$&$\alpha$=$0.2$&$\alpha$=$0.3$&$\alpha$=$0.4$&$\alpha$=$0.5$ \\
			\hline
			R-C3D+SE &13.96 &12.61 &10.81 &7.91 &5.11\\
			R-C3D+ConSE &14.16 &12.54 &10.93 &8.02 &5.29\\
			\hline
			Ours (-${TPN}_{*}$-$L_{sc}$)   &16.76 &14.76 &11.87 &9.01 &7.37\\
			Ours (+${TPN}_{*}$-$L_{sc}$)  &17.92 &15.03 &12.99 &9.61 &8.25
			\\
			Ours (-${TPN}_{*}$+$L_{sc}$)   &19.50& 16.72& 13.81&11.23 &8.88
			\\
			Ours (+${TPN}_{*}$+$L_{sc}$)   &21.34 &16.98 &15.01 &11.12 &9.15
			\\
			\hline
	\end{tabular}}
	\vspace{-0.6em}
\end{table}

\subsection{Comparative Results}
\label{experimental results}
Given this study is the first work in the novel direction of ZSTAD, there are no existing methods with which to compare our model. 
Therefore, we design two baselines called R-C3D+SE and 
R-C3D+ConSE by combining the R-C3D activity detection model with well-known ZSR frameworks SE \cite{xu2015semantic} and ConSE \cite{norouzi2013zero} together. 
Note that SE is a primary method for zero-shot action recognition, by using semantic word vector space
as the common space to embed videos and category labels.
ConSE  projects samples into a semantic embedding space via a convex combination of class label embeddings. 
%Therefore, inspired by Rahman \emph{et al.} \cite{rahman2018zero}, we design a baseline called R-C3D+ConSE by combining the R-C3D activity detection model and the ConSE ZSR method. 
%This baseline first uses the R-C3D network to generate activity proposals and calculate a distribution of scores over all seen activity labels for each proposal. 
%Then, following the ConSE strategy, the semantic correlations between the seen and unseen label embeddings are used to calculate the scores over unseen activity labels during the testing stage.
In addition, we compare four versions of our model with the designed baseline to explore its efficacy. The four versions are:
\begin{itemize}
	\vspace{-0.3 em}
	\item (-${TPN}_{*}$-$L_{sc}$) without the improved classification module in TPN subnet and without super-class clustering loss for ZSDN subnet;
	\vspace{-0.6 em}
	\item (+${TPN}_{*}$-$L_{sc}$) with the improved classification module in TPN subnet and without super-class clustering loss for ZSDN subnet;
	\vspace{-0.6 em}
	\item (-${TPN}_{*}$+$L_{sc}$) without the improved classification module in TPN subnet and with super-class clustering loss for ZSDN subnet;
	\vspace{-0.6 em}
	\item (+${TPN}_{*}$+$L_{sc}$) with the improved classification module in TPN subnet and with super-class clustering loss for ZSDN subnet.
	\vspace{-0.6 em}
\end{itemize}
%The results of the zero-shot activity detection with both datasets are analyzed as follows.
\begin{table*}[t]
	\center
	\tabcolsep=0.7pt
	\caption{Per-unseen class AP (\%) at IoU threshold $\alpha = 0.5$ on THUMOS'14 dataset. \label{tab:thumos14_2}}
	{
		\begin{tabular}{c|c|c|c|c|c|c|c|c}
			\hline
			&Baseball Pitch &Cricket Bowling &Diving &Hammer Throw &Long Jump &Shotput &Soccer Penalty &Tennis Swing \\
			\hline
			R-C3D+SE  &2.23 &3.09 &3.13 &9.21 &12.15 &3.42 &3.38 &4.29 \\
			R-C3D+ConSE  &2.21 &3.07 &3.23 &9.53 &12.54 &3.56 &3.46 &4.72 \\
			\hline
			Ours (-${TPN}_{*}$-$L_{sc}$)  &3.79 &4.03 &4.41 &14.25 &17.47 &4.98 &4.92 &5.11  \\
			Ours (+${TPN}_{*}$-$L_{sc}$) &3.92 &4.24 &4.87 &15.92 &19.78 &5.64 &5.36 &6.30 \\
			Ours (-${TPN}_{*}$+$L_{sc}$) &4.20 &4.62 &5.07 &17.02 &21.23 &6.31 &5.82 &6.78 \\
			Ours (+${TPN}_{*}$+$L_{sc}$) &4.34 &4.87 &5.03 &18.12 &20.78 &7.06 &6.03 &6.93 \\
			\hline
	\end{tabular}}
\end{table*}

\textbf{Results on THUMOS'14}: 
Table \ref{tab:thumos14_1} reports the detection performance for the eight unseen activities in terms of mean average precision (mAP) at IoU thresholds $[0.1,0.5]$ (denoted as $\alpha$). 
The average precision (AP) for each unseen class at IoU threshold $0.5$ is shown in Table \ref{tab:thumos14_2}. 
From these results, we make the following four observations: 
\begin{itemize}
	\vspace{-0.3em}
	\item All four versions of our method consistently perform better than the R-C3D+SE and R-C3D+ConSE baselines over each of the unseen class. This indicates that considering label embeddings in the ZSDN subnet is beneficial to ZSTAD task.
	\vspace{-0.6em}
	\item In most cases, the performance with ${TPN}_{*}$ is better than without. This illustrates that the new classification module in the TPN subnet, which introduces information on the background label embedding, is helpful for generating unseen activity proposals.
	\vspace{-0.6em}
	\item Comparing the results of ``+$L_{sc}$'' and ``-$L_{sc}$'', we conclude that considering the semantic clustering information of label embeddings through the super-class clustering loss can improve performance. 
	\vspace{-0.6em}
	\item With all methods, the mAP scores worsen as the value of the IoU threshold $\alpha$ increases. This is reasonable because a larger IoU threshold tends to require a more accurate boundary for the predicted activities.
\end{itemize}
\begin{table}[t]
	\center
	\tabcolsep=3.0pt
	\caption{Zero-shot temporal activity detection results on Charades w.r.t standard and post-process mAP (\%).  \label{tab:charades1}}
	{
		\begin{tabular}{c|ccccc}
			\hline
			&Standard mAP&Post-Process mAP \\
			\hline
			R-C3D+SE &5.13 &9.17\\
			R-C3D+ConSE &5.67  &9.84 \\
			\hline
			Ours (-${TPN}_{*}$-$L_{sc}$)  &6.63  &10.89\\
			Ours (+${TPN}_{*}$-$L_{sc}$) &7.03  &11.72
			\\
			Ours (-${TPN}_{*}$+$L_{sc}$) &7.57  &12.86 \\
			Ours (+${TPN}_{*}$+$L_{sc}$) &7.91  & 13.23
			\\
			\hline
	\end{tabular}}
\end{table}	

\textbf{Results on Charades}: 
Table \ref{tab:charades1} reports the results for the 37 unseen activities with the Charades data in terms of Sigurdsson \emph{et al.}'s \cite{sigurdsson2017asynchronous} standard and post-processed mAP evaluation metrics. 
The results are very consistent with those from the THUMOS’14 dataset. 
\begin{table*}[t]
	\center
	\tabcolsep=3.5 pt
	\caption{Per-unseen class AP (\%) on Charades dataset with our method (+${TPN}_{*}$+$L_{sc}$). \label{tab:charades2}}
	{
		\begin{tabular}{c|c|c|c|c|c|c|c}
			\hline
			Throwing clothes  &10.80
			& Opening a door &11.53
			& Sitting at a table &16.44
			& Talking on a phone &5.28
			\\
			\hline
			Holding a bag &7.86	
			& Taking a book &3.93
			& Reading at a book &11.66
			& Holding a towel/s &12.87
			\\
			\hline
			Taking from a box &3.58
			& Closing a box &4.08
			& Taking a laptop &3.45
			& Tidying up a blanket &5.93
			\\
			\hline
			Sitting in a chair &18.09
			& Putting food somewhere &10.94
			& Eating a sandwich &7.96
			& Taking shoes &10.88
			\\
			\hline
			Holding a pillow &7.91
			& Tidying a shelf &4.84
			& looking at a picture &5.64
			& Closing a window &3.67
			\\
			\hline
			Taking a broom &10.35
			& Holding a mirror &2.69
			& Turning off a light &4.97
			& Washing a cup&4.05
			\\
			\hline
			Opening a closet &7.54
			& Taking paper &4.11
			& Wash a dish &9.59
			& Sitting on sofa &14.41
			\\
			\hline
			Tidying on the floor &8.14
			& Holding medicine &5.04
			& Taking a vacuum &5.63
			& Lying on a bed &10.10
			\\
			\hline
			Watching television &11.12
			& Fixing a doorknob &2.87
			& Opening a refrigerator &4.50
			& Someone is eating &5.32
			\\
			\hline
			Someone is dressing &14.90
			\\	
			\hline
	\end{tabular}}
\end{table*}
That is, (+${TPN}_{*}$+$L_{sc}$) still delivers the best activity detection results with a 2.78\% and 2.24\% performance improvement in standard mAP compared to R-C3D+SE and R-C3D+ConSE, respectively. 
We also note that the manually-designed post-processing effectively improves the performance in terms of mAP. 
In addition, Table 4 shows the standard AP values per-unseen activity class with the method (+${TPN}_{*}$+$L_{sc}$). 
As the table shows, these unseen activities, such as ``Sitting in a chair'', ``Someone is dressing'', and ``Sitting at a table'', are relatively easy to detect in the testing stage.
\begin{figure}[t]
	\vspace{-1.5em}
	\centering
	\subfloat[THUMOS'14 \label{fig:thumos14_tpn}  ]{\includegraphics[width=0.53\linewidth]{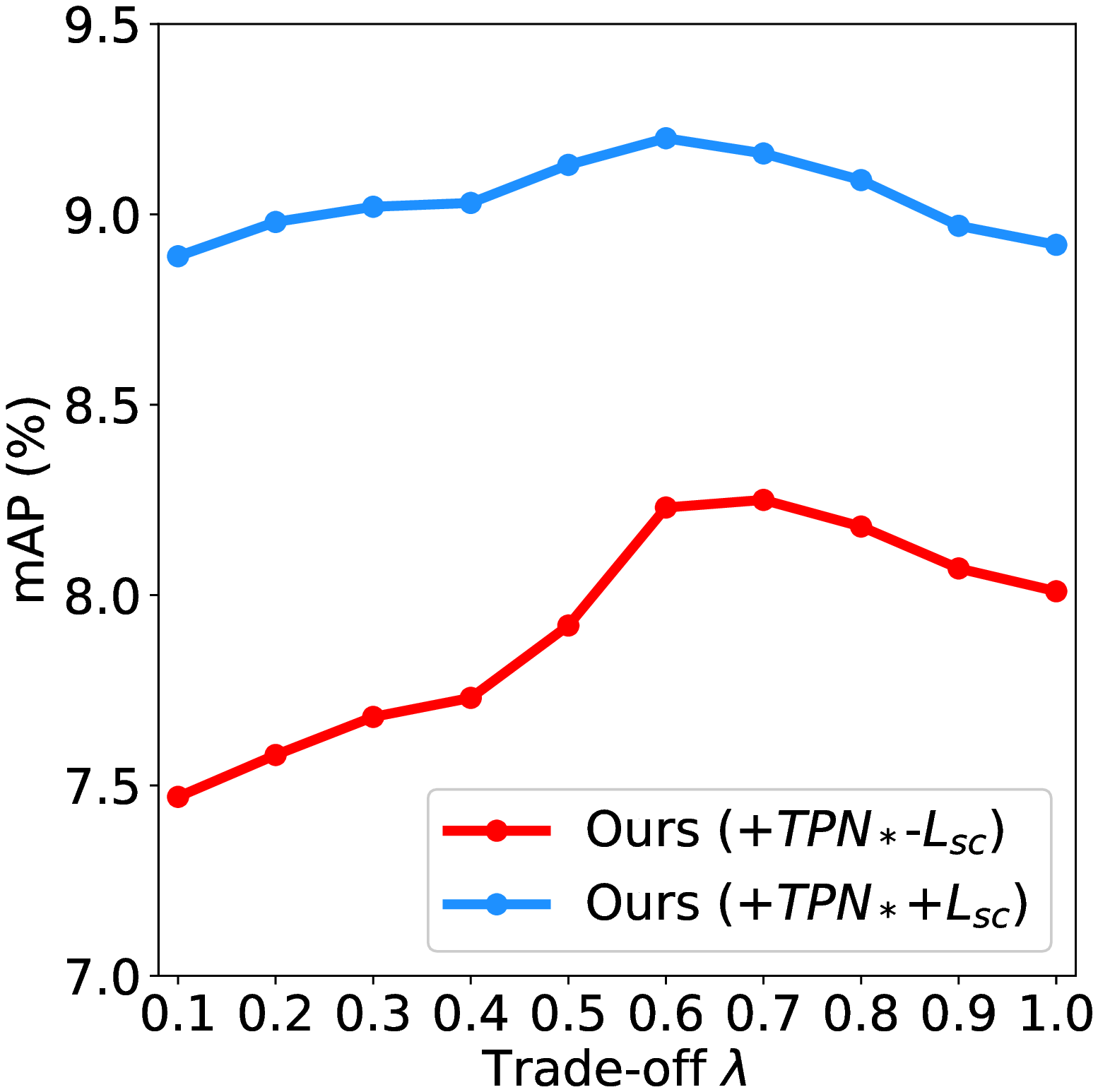}} 
	\subfloat[Charades \label{fig:charades_tpn} ]{\includegraphics[width=0.53\linewidth]{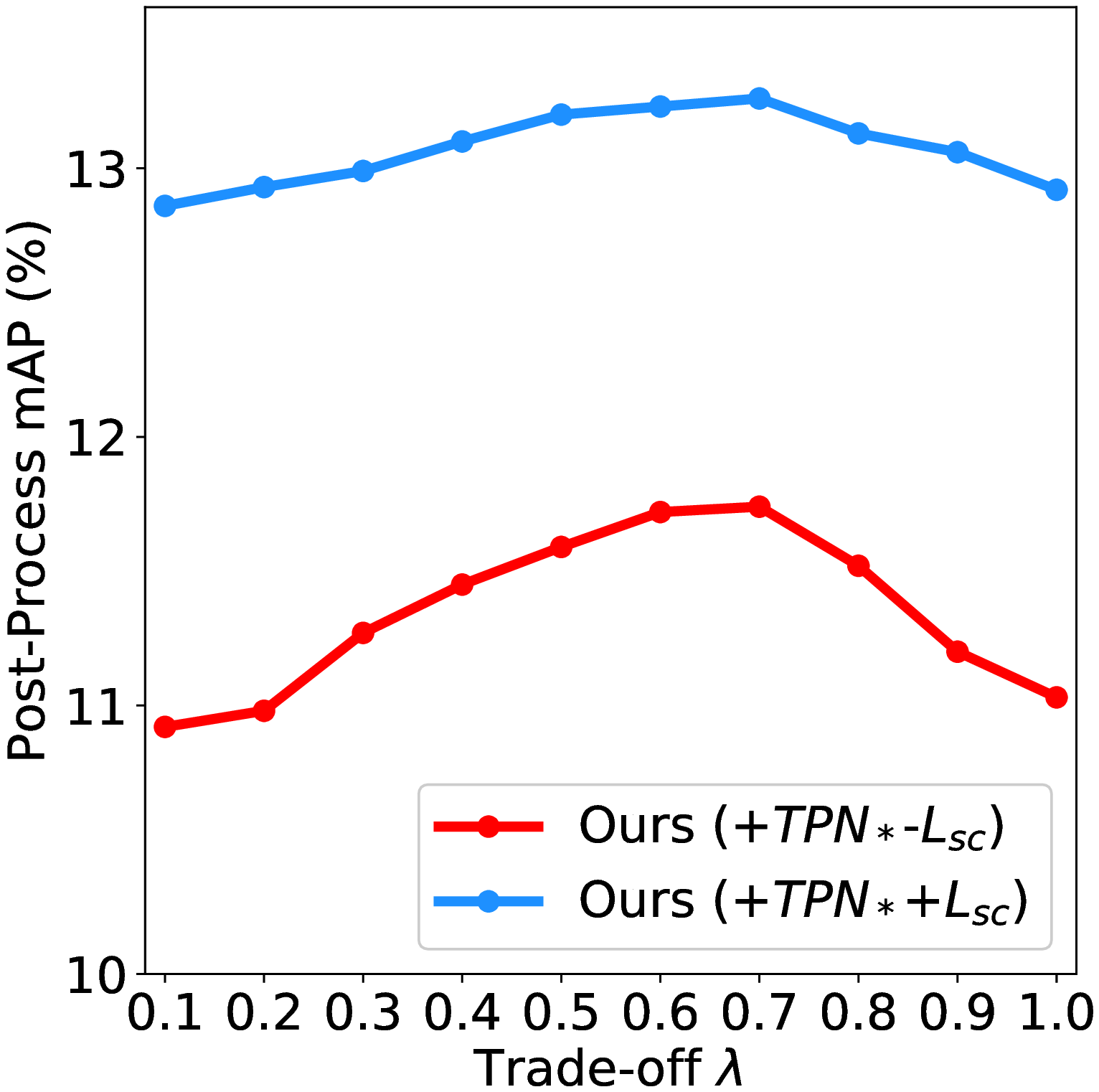}} 
	\caption{The influence of improved classification module (trade-off $\lambda$) in TPN subnet for ZSTAD. \label{fig:tpn_curves}}
\end{figure} 

\subsection{Impact of the Improved Classification Module in the TPN Subnet}
\label{Impact of Improved Classification Module in TPN}
The purpose of the improved classification module in the TPN subnet is to refine the results of previous binary classification as to whether the segment contains an activity or is the background. 
It does this by introducing the label embeddings of background and weighting the output with a trade-off hyper-parameter $\lambda$. 
Therefore, it is necessary to test the influence of $\lambda$ on the results. 
The variants (+${TPN}_{*}$-$L_{sc}$)  and (+${TPN}_{*}$+$L_{sc}$), show the effect of this parameter with $\lambda$ tuned from $0.1$ to 1 in step of $0.1$. 
Figure \ref{fig:tpn_curves} shows both the mAP curves at an IoU threshold of $\alpha=0.5$ for THUMOS’14, and the post-process mAP curves for Charades. We make two observations from the results:
\begin{itemize}
	\vspace{-0.3em}
	\item (+${TPN}_{*}$-$L_{sc}$) and (+${TPN}_{*}$+$L_{sc}$) consistently perform better than two baselines R-C3D+SE and R-C3D+ConSE  (see Tables \ref{tab:thumos14_1} and \ref{tab:charades1}) regardless of any improvement brought by the trade-off $\lambda$.
	\vspace{-0.6em}
	\item As $\lambda$ increases, performance improves to a point then gradually tapered. For THUMOS'14, the optimal value of $\lambda$ is 0.6, and 0.7 for Charades.
	\vspace{-0.6em}
\end{itemize}	
Overall, we find that the improved classification module with a properly tuned $\lambda$ enhances performance by generating better proposals for unseen activity classes in the testing stage, and the optimal value of $\lambda$ will differ depending on the specific properties of the dataset.
\begin{figure}[t]
	\vspace{-1.9em}
	\centering
	\subfloat[THUMOS'14  \label{fig:thumos14_superclasses}  ]{\includegraphics[width=0.53\linewidth]{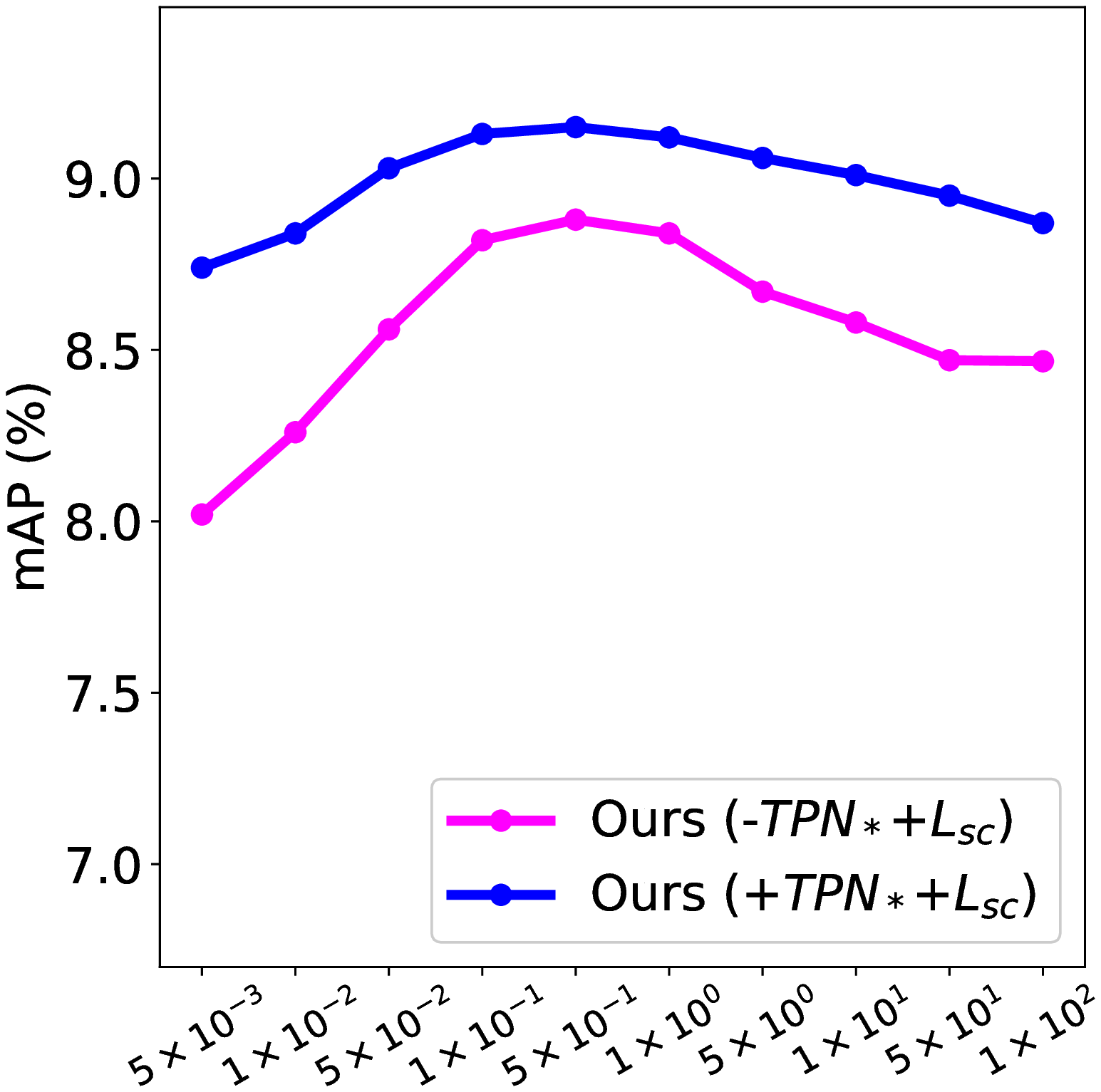}} 
	\subfloat[Charades  \label{fig:charades_superclasses} ]{\includegraphics[width=0.53\linewidth]{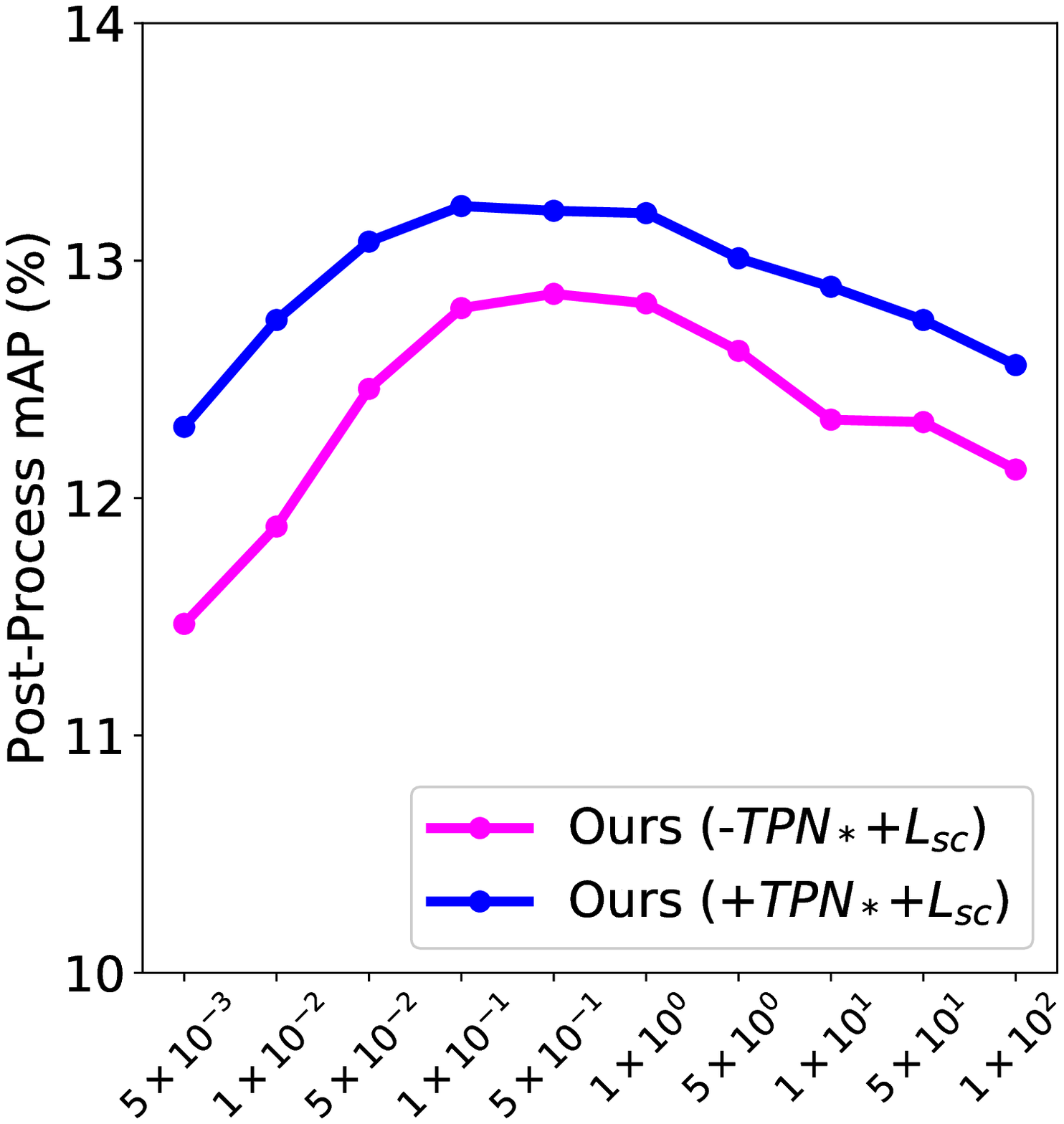}} 
	\caption{The influence of super-class clustering loss (trade-off $\beta$) in ZSDN subnet for ZSTAD. \label{fig:superclasses_curves}}
\end{figure}

\subsection{Impact of the Super-Class Clustering Loss in the ZSDN Subnet}
\label{Impact of Super-Class Clustering Loss in ZSDN}
As part of the ZSDN subnet, the super-class clustering term $L_{sc}$ in Eq. (\ref{cls_loss})  aims to capture the common semantics among all activities from the prior information of super-classes over all activity label embeddings. 
Apparently, the influence of the trade-off parameter $\beta$ in the classification loss needs to be examined. 
The variants (-${TPN}_{*}$+$L_{sc}$)  and (+${TPN}_{*}$+$L_{sc}$) are applicable here, with $\beta$ varied in interval $\{5\times 10^{-3}, 1\times 10^{-2}, 5\times 10^{-2},1\times 10^{-1}, 5\times 10^{-1},1\times 10^{0}, 5\times 10^{0},1\times 10^{1},5\times 10^{1},1\times 10^{2}\}$ for both datasets. 

Figure \ref{fig:superclasses_curves} shows the mAP curves at an IoU threshold of 0.5 for the THUMOS'14 dataset along with the post-process mAP curves for Charades. 
As the figures show, detection performance for all methods initially increases with an increase in $\beta$ before stabilizing in the interval $[10^{-1}; 10^0]$. 
Once $\beta$ exceeds $10^0$, performance gradually declines. 
In addition, we note that the results for (-${TPN}_{*}$+$L_{sc}$) are notably better than for (-${TPN}_{*}$-$L_{sc}$) regardless of the hyper-parameter $\beta$, as Tables \ref{tab:thumos14_1} and \ref{tab:charades1} show (\emph{i.e.}, a 7.37\% improvement with THUMOS’14 and a 10.89\% improvement with Charades).
Similarly, no matter the value of $\beta$, (+${TPN}_{*}$+$L_{sc}$) consistently outperforms (+${TPN}_{*}$-$L_{sc}$) on the two datasets. 
These results indicate that super-classes over activity label embeddings are conducive to mine the correlations between seen and unseen activities and significantly contribute to detecting unseen activities.

\section{Conclusion}
\label{Conclusion}	
In this paper, we propose a novel problem setting for temporal activity detection in which activities that are not seen during the training stage can be recognized and localized simultaneously. 
The solution presents in this paper is the first attempt at zero-shot temporal activity detection (ZSTAD). 
To address this challenging problem, we design an end-to-end deep network that uses label embeddings and their super-classes as prior information to capture the semantics common to seen and unseen activities. 
The results on THUMOS'14 and Charades datasets show the approach is able to detect unseen activity to a high degree of accuracy.	

\section*{Acknowledgment}
This work is supported by National Key Research and Development Program of China (2018YFB1004500), National Natural Science Foundation of China (61532004, 61532015, 61672418 and 61672419), Innovative Research Group of the National Natural Science Foundation of China (61721002), Innovation Research Team of Ministry of Education (IRT\_17R86), Project of China Knowledge Centre for Engineering Science and Technology, Intelligence Advanced Research Projects Activity (IARPA) via Department of Interior/Interior Business Center (DOI/IBC) contract number D17PC00340, Australian Research Council Discovery Early Career Researcher Award (DE190100626).

{\small
\bibliographystyle{ieee_fullname}
\bibliography{reference}
}

\end{document}